\pdfoutput=1
\documentclass[11pt]{article}

\usepackage[final]{acl}
\usepackage{times}
\usepackage{latexsym}

\usepackage[T1]{fontenc}
\usepackage[utf8]{inputenc}

\usepackage{microtype}

\usepackage{inconsolata}

\usepackage{graphicx}

\usepackage{soul}

\usepackage{tcolorbox}
\usepackage{booktabs}

\usepackage{tabularx}
\usepackage{siunitx}
\title{CLASS-IT: Conversational and Lecture-Aligned Small-Scale Instruction Tuning for BabyLMs}

\author{Luca Capone\textsuperscript{1}\thanks{Corresponding author}\thanks{For the specific purposes of Italian Academy, Luca Capone is responsible for Sections \ref{sec:related}, \ref{sec:dataset} and \ref{sec:models & train}, Alessandro Bondielli is responsible for sections \ref{sec:eval} and \ref{sec:discuss}, Alessandro Lenci is responsible for sections \ref{sec:intro} and \ref{sec:conclusions}.} \and Alessandro Bondielli\textsuperscript{1,2}\footnotemark[2] \and Alessandro Lenci\textsuperscript{1}\footnotemark[2] \\ \\
       \textsuperscript{1}CoLing Lab, Department of Philology, Literature and Linguistics, University of Pisa\\
       \textsuperscript{2}Department of Computer Science, University of Pisa\\ \\
       luca.capone@fileli.unipi.it, \{alessandro.bondielli, alessandro.lenci\}@unipi.it}
       
\begin{document}
\maketitle
\begin{abstract}
This work investigates whether small-scale LMs can benefit from instruction tuning. We compare conversational and question-answering instruction tuning datasets, applied either in a merged or sequential curriculum, using decoder-only models with 100M and 140M parameters. Evaluation spans both fine-tuning (SuperGLUE) and zero-shot (BLiMP, EWoK, WUGs, entity tracking, and psycholinguistic correlation) settings. Results show that instruction tuning yields small but consistent gains in fine-tuning scenarios, with sequential curricula outperforming merged data; however, improvements do not consistently transfer to zero-shot tasks, suggesting a trade-off between interaction-focused adaptation and broad linguistic generalization. These results highlight both the potential and the constraints of adapting human-inspired learning strategies to low-resource LMs, and point toward hybrid, curriculum-based approaches for enhancing generalization under ecological training limits.
\end{abstract}

\section{Introduction} \label{sec:intro}
The role of input data vis-à-vis innate biases has long dominated the debate on language acquisition. This is exemplified by arguments such as the poverty of the stimulus and the language of thought hypothesis \cite{chomsky1980rules, fodor1975language}, which have emphasized the need for innate constraints governing the process of acquiring productive linguistic generalizations. In contrast, data-driven learning has always been a central tenet of connectionist theory, arguing that, given sufficient training, a large enough model can reproduce any regular behavioral pattern \cite{smolensky1988proper}. One of the defining features of LMs is that performance relies on the training process. The development of model abilities clearly reflects learning, although the precise nature of this learning is not yet well understood. It remains uncertain whether abilities (or at least some of them) are truly emergent \citet{wei2022emergent}, or whether this impression is an artifact of measurement, with capabilities in fact increasing more gradually \cite{schaeffer2023emergent}. Moreover, the type and order of training data can influence a model’s ability to perform specific tasks \cite{soviany2022curriculum}. Finally, particular training regimes, such as instruction tuning or reinforcement learning with human feedback (RLHF), can significantly enhance a model’s capacity for user interaction, as well as its logical, inferential, and reasoning abilities. 
Despite relying on radically different mechanisms, LMs and humans share several key properties of learning: both improve with training over time, both are sensitive to the quality of instruction, and both benefit from interactive, feedback-driven training. These parallels suggest that current LMs approximate some aspects of human-like learning. However, the scale of resources required (both in terms of data and computation) remains orders of magnitude greater than what is needed for human learning, especially in children \cite{frank2023bridging}.
Among these shared features, this paper focuses on \textbf{interaction}, a core component of human learning, particularly in childhood. We investigate whether an LM trained on ecologically valid input, comparable in scale to the linguistic exposure of a 10-year-old child, can benefit significantly from targeted instruction tuning. Specifically, we compare two types of instruction tuning datasets: one centered on conversational interactions and the other focused on question-answering tasks.
The main research questions addressed in this study are:
\begin{itemize}
\item Can a BabyLM benefit from instruction tuning?
\item Given the limited pre-training typical of BabyLMs, which type of instruction data is more effective: conversational or open-ended question-answering? 
\item Does a curriculum learning approach to instruction tuning provide significant benefits?
\end{itemize}

This paper is organized as follows. Section~\ref{sec:related} reviews related work. Section~\ref{sec:dataset} describes the datasets used for pre-training and instruction tuning. Section~\ref{sec:models & train} presents the model architectures and details the training procedures, while Sections~\ref{sec:eval} and~\ref{sec:discuss} present and analyze the results on the BabyLM Challenge tasks. Finally, Section~\ref{sec:conclusions} summarizes our findings and outlines directions for future research.

\section{Related Works}\label{sec:related}
While early language learning in children is often portrayed as remarkably precocious \cite{mccormack2005children, gopnik2011theory, dundar2020children}, linguistic and psychological studies suggest that this view must be qualified. Many scholars acknowledge children’s early communicative abilities, but argue that these are constrained to specific tasks and contexts, and do not necessarily reflect a fully developed understanding of language. For instance, although the intersubjective (i.e., social and communicative) function of linguistic signs becomes evident in children from an early age, their perspectival function, the ability to conceptualize experiences from multiple viewpoints, emerges more gradually \cite{vygotsky1987collected,piaget2002judgement,tomasello2009cultural}. 
%\cite[p. 123]{tomasello2009cultural}. 

%Piaget noted that many children up to the age of seven \cite[p. 17]{piaget2002judgement}, despite producing fluent and age-appropriate speech, struggled to complete sentences involving terms such as \textit{because}, \textit{therefor}, or \textit{although}, that require representing events from a logical-causal or anthitetical perspective \cite[p. 19]{piaget2002judgement}.
%Vygotsky reported similar findings. He compared children’s use of spontaneous concepts (concepts derived from everyday experience) with scientific concepts (acquired through formal instruction). The study found that scientific concepts showed more sophisticated reasoning, while spontaneous concepts remained less developed \cite[p. 279]{vygotsky1987collected}.

Drawing on developmental psycholinguistic evidence \cite{berman2013relating, peterson1987connective}, \citet{tomasello2003constructing} observed that many children up to the age of nine, despite producing fluent, age-appropriate speech, struggle to use sophisticated conjunctions (such as \textit{because, indeed, although}, etc.) when required to do so. These conjunctions involve representing events from a logical–causal or antithetical perspective, which can pose significant challenges.
At this stage, \textit{and} remains the most frequently used connective, functioning in an undifferentiated way to express a wide range of semantic relations, even after more specific connectives have begun to appear in a child’s speech.
Similar limitations occur with other complex constructions: comprehension and voluntary use often do not match the apparent fluency of spontaneous speech. \citet{berman2013relating} document the difficulties children face with narrative discourse, sometimes even up to age nine, when asked to describe a story depicted in a sequence of images. Children frequently struggle to produce coherent narratives that clearly indicate a beginning, progression, and conclusion.
\citet{tomasello2003constructing} attributes these challenges to the \textit{plurifunctionality} of complex constructions, arguing that mastery of the perspectives they encode develops gradually over the course of the school years.

%According to these scholars, linguistic forms encode particular perspectives through which events, objects, and relations are represented, perspectives not readily accessible without guidance. Tomasello argues that although spontaneous child speech may suggest cognitive depth, a closer analysis reveals that children do not initially master this perspectival dimension. Instead, it develops gradually through interactive exposure, where children learn how words and constructions can be used to structure and convey their experiences in socially shared meaningful ways \cite[p. 120]{tomasello2009cultural}.

Building on this body of research, the present study investigates whether and to what extent interactive instruction can enhance the training of a BabyLMs. In particular, it examines whether formal, instruction-like input provides greater benefits than conversational data for fine-tuning LMs trained on limited, child-comparable linguistic exposure.
To our knowledge, the two main attempts to interactively train BabyLMs using more pedagogically structured data are Baby's CoThought \cite{zhang2023baby} and Baby Stories \cite{zhao2023babystories}. However, both differ from the approach proposed in this work, albeit for different reasons. \citet{zhang2023baby} build an educational dataset based on the BabyLM Challenge trainset, using GPT-3.5-Turbo. However, the dataset is used to train an encoder-only model through  masked language modeling. 
\citet{zhao2023babystories}, on the other hand, preserves an interactive setup by fine-tuning a decoder model using proximal policy optimization. Nonetheless, this training technique departs from the type of formal instruction we aim to address, as the model is simply optimized to prefer certain generations based on a reward model. In contrast, the instruction tuning proposed in this work more closely resembles the structured, formal education typically provided to children in school settings. The present work instead adopts an instruction fine-tuning approach, where models are explicitly trained to respond to questions about specific topics and to provide appropriate answers within conversational contexts.

\section{Dataset}\label{sec:dataset}
The dataset used for model pre-training is a curated subset of the data provided by the task organizers, which amounts to approximately 91 million words. The instruction tuning dataset includes processed Switchboard transcripts and augmented Simple Wikipedia texts, enhanced using the \texttt{LLaMA-3.2-3B-Instruct} model \cite{dubey2024llama}.

\subsection{Pretraining Dataset}
The data supplied by the organizers (approximately 100 million words) underwent standard preprocessing. Special characters were removed, and all entries containing two words or fewer were discarded. Additional processing was applied to the Switchboard corpus, utterances from the same speaker were concatenated when they occurred in sequence, following standard dialogue normalization practices. Roughly 75 million words—drawn from CHILDES, Gutenberg, BNC and OpenSubtitles—were used exclusively for pre-training. An additional 16 million words (from Switchboard and Simple Wikipedia) overlap with the instruction tuning dataset, bringing the total pre-training corpus to approximately 91 million words.

%\begin{figure*}[!h]
%\includegraphics[scale=0.6]{latex/figures/%pt_distribution_new.png}
%\caption{}
%\label{pt_distribution}
%\end{figure*}

\subsection{Instruction-Tuning Dataset}

The instruction tuning dataset consists of two sections: a \textbf{conversational component} based on the Switchboard corpus and an \textbf{instructional component} based on Simple Wikipedia. For the conversational section, the Switchboard data were adapted to meet the requirements of instruction tuning training task. Consecutive utterances from the same speaker were merged to ensure a consistent alternation between speakers' turns (e.g., A, B, A, B). The dialogues were then segmented into prompt–reply pairs using a sliding window approach with the following schema: (A1, B1), (B1, A2), (A2, B2). The resulting dataset contains 38,802 items and approximately 1.3 million words (excluding prompt-reply duplicates).
For the instructional section, Simple Wikipedia data were augmented using \texttt{LLaMA-3.2-3B-Instruct} \cite{dubey2024llama}. For each article text, three question–answer pairs were generated using structured generation with \emph{outlines}\footnote{https://dottxt-ai.github.io/outlines/latest/\#acknowledgements} and the following prompt:

\begin{tcolorbox}
Based on the following text, generate 3 questions and detailed, informative answers.
Each answer should be easy for a young person to understand and at least 2–3 sentences long.
Explain things in simple language, with clear and friendly sentences.
Avoid short or vague replies and give enough detail so a kid can learn something new.
\end{tcolorbox}

\begin{figure*}[!t]
\includegraphics[scale=0.9]{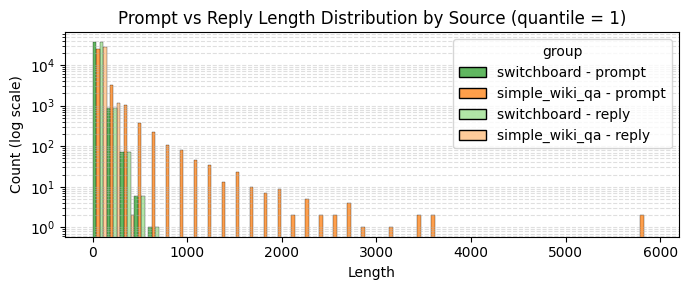}
\caption{}
\label{it_distribution}
\end{figure*}

The generated data significantly exceeds the word limit imposed by the challenge. In this work we use only a representative portion of the whole dataset (\href{https://huggingface.co/datasets/colinglab/CLASS_IT}{\texttt{colinglab/CLASS\_IT}}). The full dataset will be released in the future following appropriate validation. The subset used in this study contains 8.7 million words, keeping the total—along with the 91 million pre-training words (which already include Switchboard and Simple Wikipedia texts)—within the 100-million-word limit. The augmented Simple Wikipedia dataset includes 97,697 items, totalling 18 million words (Figure \ref{it_distribution}).

\section{Models and training}\label{sec:models & train}
%-baselines:
%BabyLM-community/babylm-interaction-baseline-simpo
%BabyLM-community/babylm-baseline-100m-gpt2
%BabyLM-community/babylm-baseline-100m-gpt-bert-mixed

%-nostri:
%140M ca. %sftp://131.114.51.106/extra/luca.capone/babylmchallange25/eval_ready_models/llama117M_pt
%100M ca. %sftp://131.114.51.106/extra/luca.capone/babylmchallange25/eval_ready_models/llama100M_pt

We trained two models (Table \ref{tab:architectures}), both based on decoder-only, LLaMA-style architectures with large maximum sequence lengths to accommodate the long texts present in the instruction tuning dataset. The first model has 140 million parameters, featuring a larger hidden size and vocabulary size. Following its training, we developed a second model with approximately 100 million parameters, using a reduced hidden size and a vocabulary size comparable to baseline models.

\begin{table}[t]
\centering

\setlength{\tabcolsep}{4pt}
\begin{tabular}{@{}lrr@{}}
\toprule
\textbf{Hyperparameter} & \textbf{llama140M} & \textbf{llama100M} \\ 
\midrule
Vocab size              & 32{,}000      & 16{,}384 \\
Max length              & 6{,}144       & 6{,}000  \\
Hidden size             & 704           & 512     \\
Attention heads         & 11            & 8       \\
Layers                  & 12            & 20      \\
Trainable parameters    & 140{,}231{,}872  & 100{,}684{,}288 \\ 
\bottomrule
\end{tabular}
\caption{Model architectures}
\label{tab:architectures}
\end{table}

%\begin{table}[!h]
%\centering
%\begin{tabular}{@{}lrr@{}}
%\toprule
%\textbf{Hyperparameter}         & \textbf{llama140M}    & \textbf{llama100M} \\ \midrule
%Vocab size                      & 32.000                 & 16.384  \\
%Max length                      & 6.144                    & 6.000\\
%Hidden size                     & 704                    & 512\\
%Attention heads                 & 11                    & 8\\
%Layers                          & 12                    & 20\\
%Trainable parameters            & 140.231.872                     & %100.684.288\\ \bottomrule
%\end{tabular}
%\caption{Model architectures}
%\label{tab:architectures}
%\end{table}

\begin{table}[t]
\centering
\setlength{\tabcolsep}{4pt}
\begin{tabular}{@{}lrr@{}}
\toprule
\textbf{Hyperparameter} & \textbf{Pretrain} & \textbf{Instr.\ tuning} \\ 
\midrule
Initial LR              & 2e-4   & 2e-5 \\
Batch size              & 8      & 8 \\
Maximum epochs          & 8      & 10 \\
LR scheduler            & linear & cosine w/ restarts \\
Warm-up steps           & 5{,}000 & 500 \\ 
\bottomrule
\end{tabular}
\caption{Training parameters}
\label{tab:train-param}
\end{table}

%\begin{table}[!h]
%\centering
%\begin{tabular}{@{}lrr@{}}
%\toprule
%\textbf{Hyperparameter}         & \textbf{Pretrain} & \textbf{Instruction-tuning} \\ \midrule
%Initial LR           & 2e-4              & 2e-5   \\
%Batch size                      & 8                 & 8 \\
%Maximum epochs                  & 8                 & 10 \\
%LR scheduler                    & linear            & cosine with restarts \\
%Warm-up steps                   & 5.000              & 500 \\ \bottomrule
%\end{tabular}
%\caption{Training parameters}
%\label{tab:train-param}
%\end{table}

%\section{Training}\label{sec:training}
Both models followed the same pre-training procedure. The tokenizer was trained on the entire available corpus before pre-training began. Models were then pre-trained for 8 epochs using the parameters in Table \ref{tab:train-param}, processing a total of approximately 728 million words.

Instruction tuning use a different set of hyperparameters from pre-training, but the same instruction tuning configuration was applied to both models (see Table \ref{tab:train-param}). Each model was fine-tuned for 10 epochs, processing an additional 180 million words.
In total, each model processed around 908 million words across both pre-training and instruction tuning. All datasets were split 90/10 into training and validation sets, with only the training portion contributing to parameter updates. Consequently, roughly 90\% of the 908 million words (about 817 million) directly influenced model weight updates. We used the same token-level cross-entropy loss used for pre-training. However, for instruction tuning, we compute the loss only on target tokens, e.g. the answer tokens in a question-answer data point.

We adopted two strategies for instruction tuning: \textbf{merged} and \textbf{sequential}. In the merged strategy, augmented Simple Wikipedia data were shuffled together with Switchboard data, mixing conversational and instructional items. This produced the \texttt{it\_merged} models (see Figure \ref{finetune_results}). In the sequential strategy, the two datasets were used in succession, resulting in two variants: \texttt{it\_switch\_wiki} and \texttt{it\_wiki\_switch}, depending on the order in which the pre-trained model was exposed to the instruction tuning datasets. This approach was designed to test whether keeping the tasks separate—and whether the order of exposure—provides measurable benefits to model performance.

\section{Evaluation and Results}\label{sec:eval}

To evaluate our models, we used the official data provided by the challenge organizers. The evaluation is distinguished between a fine-tuning evaluation and a zero-shot evaluation.

\begin{figure*}[!h]
\includegraphics[width=\textwidth]{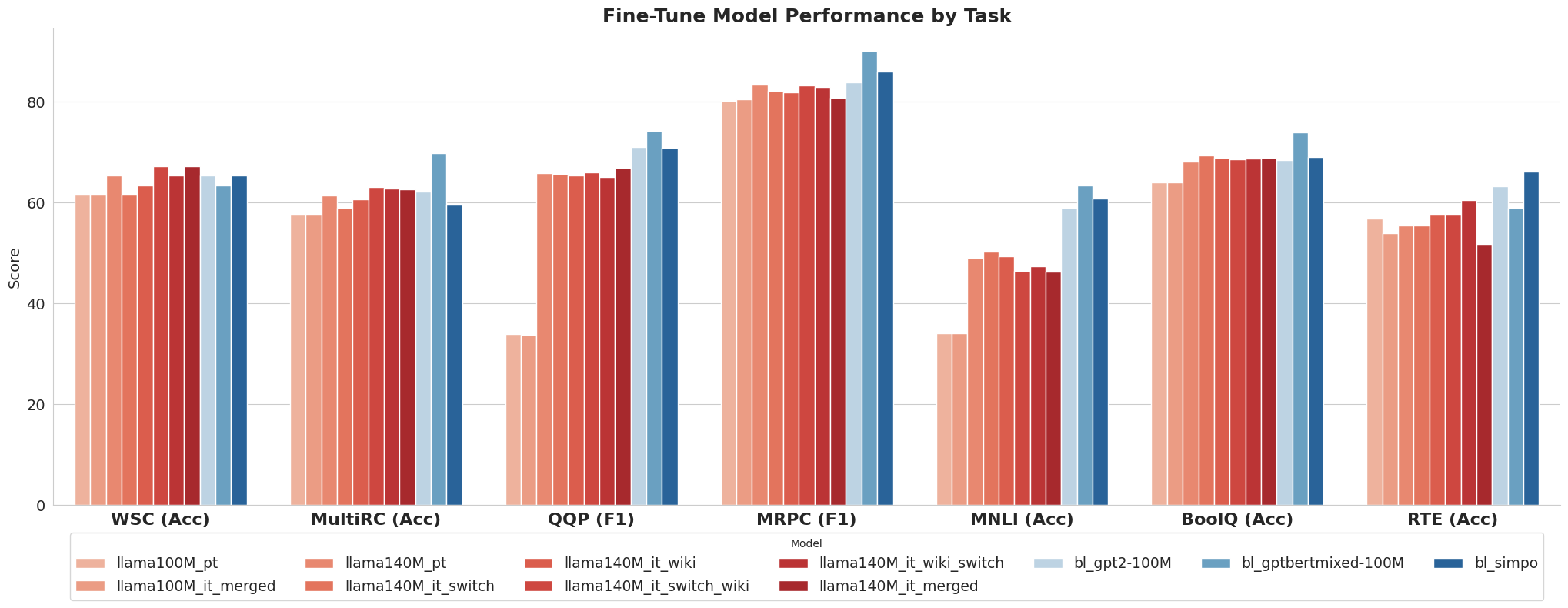}
\caption{Results of fine-tuned models on (Super)Glue tasks.}
\label{finetune_results}
\end{figure*}

\paragraph{Fine-Tuning Evaluation.} In the fine-tuning evaluation, the models are fine-tuned and evaluated in the (Super)GLUE \cite{sarlin2020superglue} tasks. We leave all the default parameters unchanged during training on each task. Note that the fine-tuning dataset is composed of a randomly sampled 10k portion of the original training set for the task. Models are evaluated on the test set. 
Figure \ref{finetune_results} shows the result of our models (both pre-trained and instruction-tuned) and the baselines (\texttt{bl\_gpt2-100M, bl\_gptbertmixed-100M, bl\_simpo}, shown in blue).

We observe that our models are generally competitive with the baselines, albeit surpassing them only for some configurations in the WSC task. As for the model size, the 140 million parameter models are generally better than the 100 million ones. However, the only tasks where the difference is noticeable are QQP and MNLI. Here, the 100 million models are markedly worse than their 140 million siblings, and both are markedly worse than the baselines.
As for the instruction fine-tuning, it seems relatively beneficial. We see in fact that in all cases there is at least an instruction-tuned model better than the pre-trained one. However, differences are small and inconsistent across tasks, that is, there is no instruction tuning configuration that systematically leads to better results.

\begin{figure}[!h]
\includegraphics[width = \linewidth]{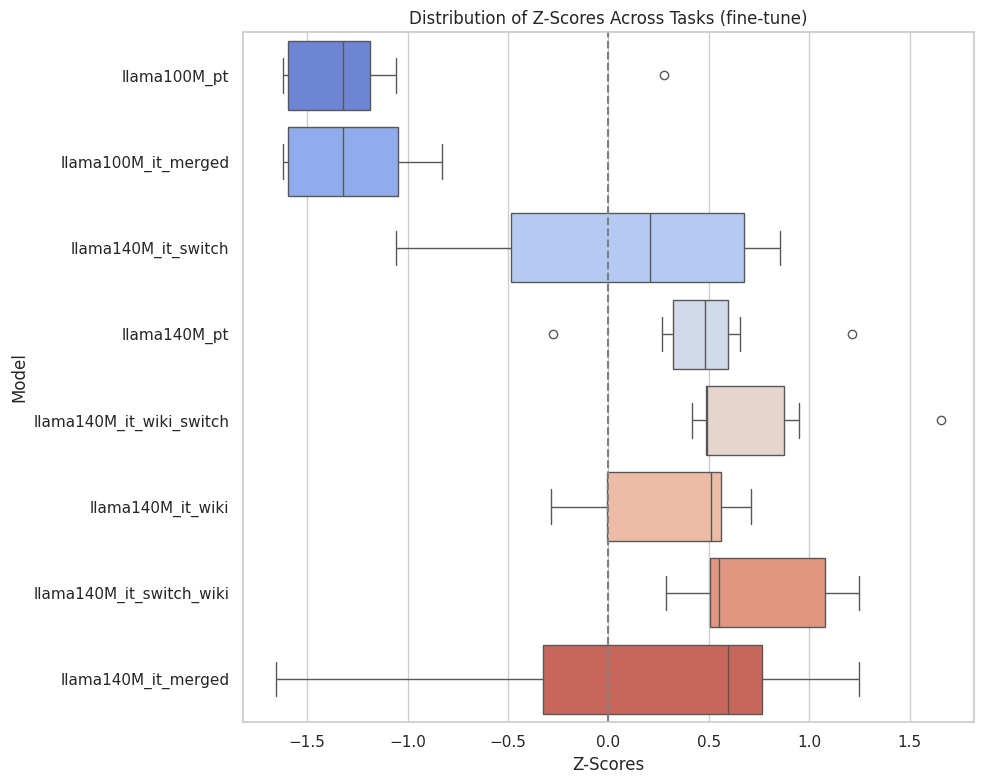}
\caption{Median, Inter-Quartile Ranges (IQR), and outliers for z-scores of each model in the \textbf{fine-tuning} evaluation.}
\label{finetune_zscore_results}
\end{figure}

Since we have multiple tasks and models, we needed a way to compare performance globally. To achieve this, we standardized the results by computing z-scores for each model on each task, which express how many standard deviations above or below the task mean a model’s score lies. We then averaged these z-scores across tasks to obtain a single global index per model. This index reflects the overall relative standing of a model compared to others, rather than absolute task performance, and allows fair comparison across heterogeneous metrics.
Specifically, we plot them including median, Inter-Quartile Ranges (IQR), and outliers. Results are in Figure \ref{finetune_zscore_results}.

We observe that the 100 million models, both pre-trained and instruction-tuned, have negative z-score medians, while all the 140 million variants are on the positive side of the plot, showcasing that differences between the smaller and larger models appear to be significant. 
Regarding the differences between pre-training only and instruction tuning, we notice some interesting aspects. The model with the highest median is the merged instruction-tuned variant trained on a mixture of the datasets. However, both models trained on the two dataset sequentially, regardless of the order, have a very similar median score, but a much smaller IQR, and are the only two models with all z-scores above zero. Variants trained only on one of the datasets perform worse, and on par with the pre-trained only model, with the one trained just on Switchboard performing worst.

\paragraph{Zero-shot Evaluation.} In the zero-shot scenario, models are evaluated using log-probabilities of sequences and/or words to obtain either model predictions or compute correlations with human data. The zero-shot evaluation is conducted on the following datasets: BLiMP \cite{warstadt2020blimp} and EWoK \cite{ivanova2024elements} are standard minimal pairs datasets that test linguistic and world knowledge of LLMs and were included also in previous years' evaluations; a WUGs task designed to understand abilities in adjective nominalization \cite{hofmann2025derivational}; an entity tracking task on data from \cite{kim2023entity}; a correlation evaluation where cloze probability, predictability ratings, and computational estimates are compared against EEG and human reading time data \cite{de2024cloze}.

\begin{figure*}[!h]
\centering
\includegraphics[width=\textwidth]{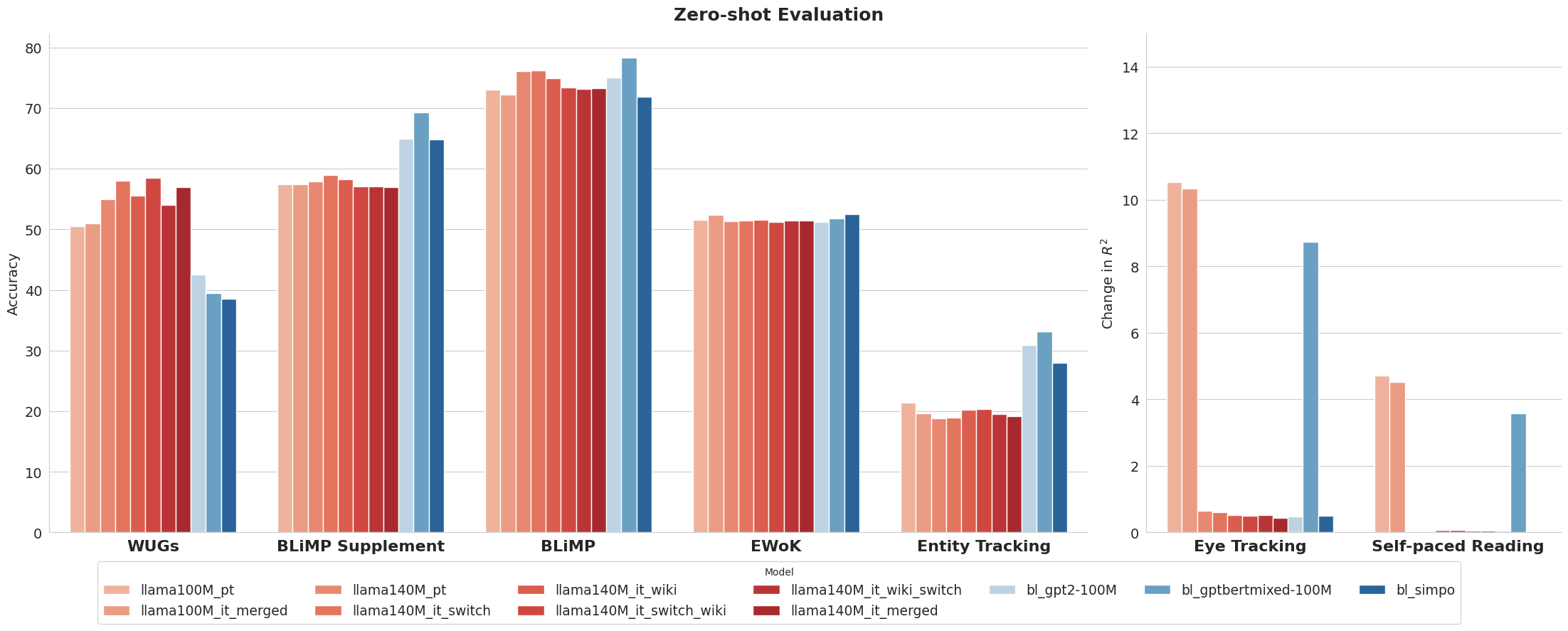}
\caption{Results of the zero-shot evaluation. Tasks measured with accuracy are reported in the left bar chart; tasks measured with change in $R^2$ are reported in the bar chart on the right. }
\label{zeroshot_results}
\end{figure*}

\begin{figure}[!h]
\includegraphics[width=\linewidth]{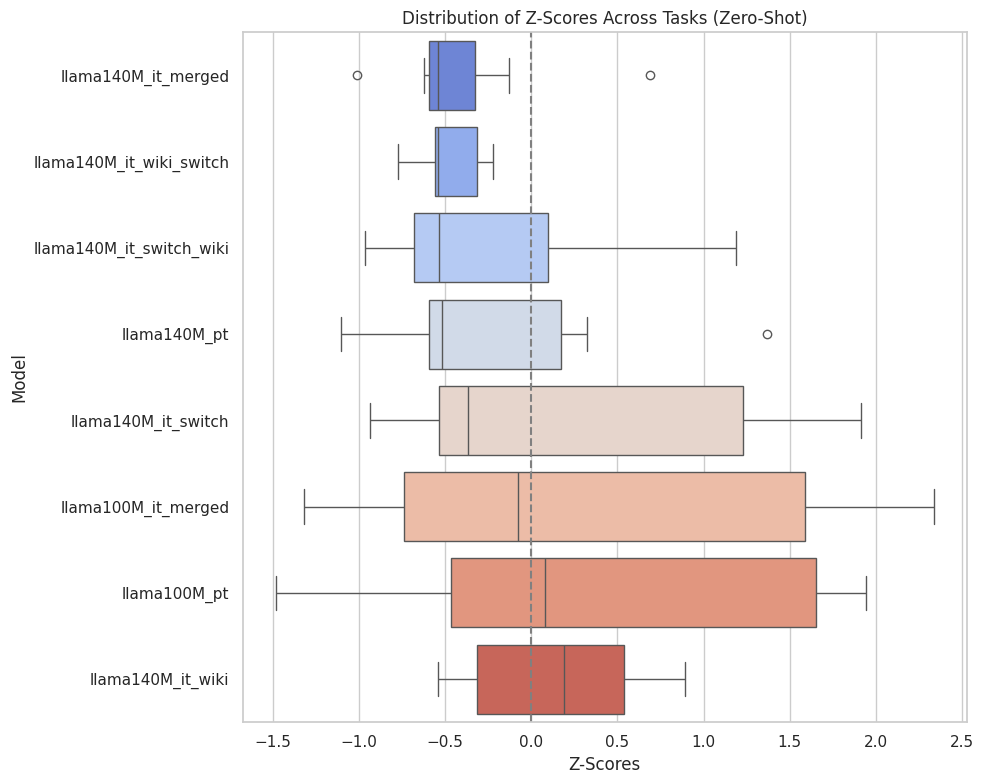}
\caption{Median, Inter-Quartile Ranges (IQR), and outliers for z-scores of each model in the \textbf{zero-shot} evaluation.}
\label{zeroshot_zscore_results}
\end{figure}

Results are reported in Figure \ref{zeroshot_results}. For the accuracy-based tasks, we do not observe striking differences between pre-trained and instruction-tuned models, similarly to what seen in the fine-tuning evaluation. However, here we also do not observe large differences also between the 100 million and 140 million variants. Our models seem to vastly outperform baselines on the WUGs task, but are worse on the Entity Tracking task, on which all models including baselines seem to struggle. As for the Change in $R^2$ based tasks, we observe some surprising results: The 100 million model variants are vastly superior to both the 140 million models and the baselines, which score almost zero with the exception of the \texttt{GPT-BERT} mixed model. We compute z-scores distribution also in this case, and report them in Figure \ref{zeroshot_zscore_results}.

For the zero-shot evaluation the z-score distribution is radically different. No model has all z-scores above zero, and only two of them has a median z-score above zero. The two 100 million variants are among the best performing models, albeit this could be attributed to the vast differences between them and all the other models on the $R^2$-based tasks.
The best performing model is an instruction-tuned variant, specifically the one trained only on Simple Wikipedia. However, no clear trend in favour or against instruction tuning emerge from the plot.

\begin{figure}[!h]
\includegraphics[width=\linewidth]{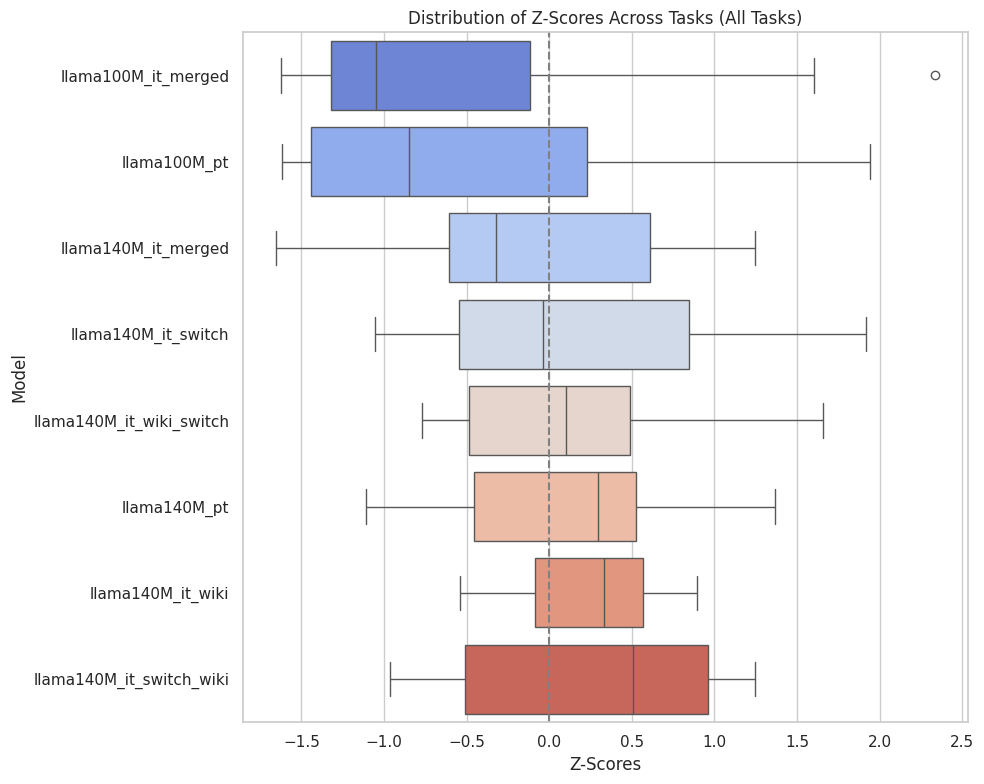}
\caption{Median, Inter-Quartile Ranges (IQR), and outliers for z-scores of each model including both zero-shot and fine-tuning evaluations.}
\label{all_zscore_results}
\end{figure}

In order to further examine the performances on a broader level, we also plot the z-score distribution including both zero-shot and fine-tuning evaluations. Results are shown in Figure \ref{all_zscore_results}. It highlights the fact that, overall, the larger models seem to perform better. As for the impact of instruction tuning, we can highlight three aspects. First, we see that the best overall model is an instruction-tuned one. However, we cannot extrapolate a clear trend in favour of instruction tuning. Second, we observe that tuning the model sequentially on different datasets is consistently better than doing so on a mixture of the datasets. The order of the instruction tuning task seems less relevant, albeit we see that tuning first on conversational data (Switchboard) and then on question answering (Simple Wiki) seem to yield better results. This however may be affected by the difference in size between the datasets. In fact, we see that the model trained only on question answering performs better than the one trained subsequently on conversations. 

\section{Discussion}\label{sec:discuss}

Our experiments provide some interesting insights about small-scale instruction tuning models trained on ecological amounts of data. 

First, we see that instruction tuning appears to be somewhat beneficial, especially if the model is further fine-tuned on specific tasks; the same improvement are not as apparent on the zero-shot evaluation. We can hypothesize that the instruction tuning stage varies the models' internal distribution to a higher degree, especially at this scale, thus affecting the performances on zero-shot tasks, where the encoding of grammar rules (BLiMP, WUGs) or specific facts (EWoK) is more relevant than conversational and/or generative performances, which are not tested here. 
The instruction tuned models may be biased to learn to solve a specific task, in our case following a conversation or answering factual questions, thus losing their generalization abilities on just language. This latter aspect is quite interesting in the context of BabyLMs, as larger models have been shown to not suffer from similar issues \cite{miliani2025explica}. Further evidence for this effect can be seen in the fact that instruction-tuned models have seen more data than pre-trained ones, yet do not consistently outperform them.

Second, we observe that among our models, smaller ones are consistently better than larger ones at correlating with human data, with the pre-trained model being slightly better; only one baseline model, with a different architecture but of the same size as ours, achieve comparable results. This is in line with previous literature where smaller models often correlate better with human psychometric data \cite{de2023scaling, oh2023does}.

Finally, we see that all our models achieve relative consistent performance on a single-task basis, while being very inconsistent across tasks and evaluation methods; the same happens with baselines. This suggests that the constraints posed by the challenge itself, namely the amount of data and training compute allowed for a training run, may limit the generalization capability of decoder-only style models without additional modifications. 

\section{Conclusion and future work}\label{sec:conclusions}
This study examined whether BabyLM-scale models (trained on ecologically realistic amounts of linguistic input) can benefit from instruction tuning, and how different forms and orders of data affect their performance. Our findings show that instruction tuning yields modest but measurable gains in fine-tuning scenarios, particularly when conversational and question–answering datasets are presented sequentially rather than merged. However, these benefits do not translate consistently to zero-shot evaluations, suggesting that, at this scale, instruction tuning may bias models toward narrow interactional behaviors at the expense of broader linguistic generalization.

A further limitation lies in how models are evaluated in the Challenge. In fact, in the fine-tuning evaluation most tasks are actually classification tasks, on which masked LMs may prove more reliable; in the zero-shot task, the vast majority of evaluations are conducted using log-likelihood as proxy for model choices. While this choice is valid in the context of the challenge, to accomodate the largest possible number of architectures and simplify the evaluation process, we can argue that model performances, especially when considering conversational instruction tuning, may be undermined by the evaluation criteria. 
Moreover, the chosen conversational portion of the instruction tuning dataset may limit the performances of the model: while the Switchboard corpus offers a structured and well-annotated resource, it represents a restricted register of spoken English and lacks much of the contextual diversity found in everyday interaction. More ecologically valid conversational data, spanning a wider range of speakers, settings, and discourse types, would provide a richer foundation for model adaptation and a stronger basis for subsequent instructional fine-tuning, potentially improving both interactive competence and generalization.
Notably, smaller models exhibited stronger correlations with human psycholinguistic data, echoing prior observations that reduced capacity can sometimes yield representations more aligned with human processing patterns. Overall, the results highlight both the promise and the limitations of adapting human-inspired learning strategies to small-scale LMs: interaction helps, but the gains are context-dependent, and generalization remains challenging under strict data and compute constraints. Future work should explore hybrid approaches that combine instruction tuning with targeted multi-task or curriculum learning, investigate architectures better suited for low-resource generalization, and extend the evaluation to interactive and communicative benchmarks that more directly reflect the ecological learning conditions motivating the BabyLM challenge.

\section*{Limitations}

Our instruction tuning experiments are constrained by the relatively small size of the instruction tuning datasets compared to pre-training corpus, which may have reduced the impact of instruction-specific learning. A different allocation (using more instruction tuning data and proportionally less pre-training data) might yield stronger effects. Moreover, the balance between question–answering and conversational data is imperfect, with the latter under-represented, potentially biasing results toward factual over interactive skills. Finally, the Simple Wikipedia augmentation process was only partially validated, and higher-quality or more diverse instructional sources could improve both robustness and generalization.

\section*{Acknowledgments}

We acknowledge financial support under the PRIN 2022 Project Title "Computational and linguistic benchmarks for the study of verb argument structure" – CUP I53D23004050006 - Grant Assignment Decree No. 1016
adopted on 07/07/2023 by the Italian Ministry of University and Research (MUR). This work was also supported under the  PNRR—M4C2—Investimento 1.3, Partenariato Esteso PE00000013—“FAIR—Future Artificial Intelligence Research”—Spoke 1 “Human-centered AI,” funded by the European Commission under the NextGeneration EU programme''

%\bibliographystyle{acl_natbib}
\bibliography{custom}

\appendix

%\section{Example Appendix}
%\label{sec:appendix}

%This is an appendix.

\end{document}